\documentclass[sigconf]{acmart}

\AtBeginDocument{%
  \providecommand\BibTeX{{%
    \normalfont B\kern-0.5em{\scshape i\kern-0.25em b}\kern-0.8em\TeX}}}

\usepackage{soul}

\usepackage{multirow}
\usepackage{enumitem}
\usepackage{microtype}
\usepackage{hyperref}

\usepackage{subcaption}

\newcommand\BibTeX{B\textsc{ib}\TeX}
\newcommand{\stitle}[1]{\noindent{\textbf{#1}}}

\DeclareMathAlphabet{\mathpzc}{OT1}{pzc}{m}{it}
\makeatletter
\newcommand*\bigcdot{\mathpalette\bigcdot@{.5}}
\newcommand*\bigcdot@[2]{\mathbin{\vcenter{\hbox{\scalebox{#2}{$\m@th#1\bullet$}}}}}
\makeatother

\newcommand{\one} {\mathpzc{1} }
\newcommand{\two} {\mathpzc{2} }

\newcommand{\bigS} {\mathcal{S} }
\newcommand{\bigX} {\mathcal{X} }
\newcommand{\bigY} {\mathcal{Y} }

\newcommand{\bigK} {K}
\newcommand{\bigU} {\mathcal{U}}
\newcommand{\bigH} {\mathcal{H} }
\newcommand{\bigD} {\mathcal{D} }
\newcommand{\bigC} {\mathcal{C} }

\newcommand{\bigP} {\mathcal{P} }
\newcommand{\bigQ} {\mathcal{Q} }

\newcommand{\smalli} {\mathpzc{i} }

\newcommand{\smalls} {s}
\newcommand{\smalll} {l}

\newcommand{\smalln} {n}
\newcommand{\smallj} {j}
\newcommand{\smallk} {k}

\newcommand{\smallt} {t}

\newcommand{\smallb} {b}
\newcommand{\smallm} {m}
\newcommand{\smallq} {q}

\newcommand{\smallr} {r}

\newcommand{\realR} {\mathbb{R} }

\newcommand{\ptod} {\mathsf{P}\mbox{-}\mathsf{ToD}}
\newcommand{\ourmodel}{$\mathsf{P}\mbox{-}\mathsf{ToD}$}

\newcommand{\myspecial}[1] {\texttt{#1}}

\newcommand{\myNum}[1]{(\emph{#1})}

\copyrightyear{2022}
\acmYear{2022}
\setcopyright{acmcopyright}
\acmConference[CIKM '22]{Proceedings of the 31st ACM International Conference on Information and Knowledge Management}{October 17--21, 2022}{Atlanta, GA, USA}
\acmBooktitle{Proceedings of the 31st ACM International Conference on Information and Knowledge Management (CIKM '22), October 17--21, 2022, Atlanta, GA, USA}
\acmPrice{15.00}
\acmDOI{10.1145/3511808.3557417}
\acmISBN{978-1-4503-9236-5/22/10}

\begin{document}

%%
%% The "title" command has an optional parameter,
%% allowing the author to define a "short title" to be used in page headers.
\title[Personalizing Task-oriented Dialog Systems via Zero-shot Generalizable Reward Function]{Personalizing Task-oriented Dialog Systems via Zero-shot Generalizable Reward Function}

%%
%% The "author" command and its associated commands are used to define
%% the authors and their affiliations.
%% Of note is the shared affiliation of the first two authors, and the
%% "authornote" and "authornotemark" commands
%% used to denote shared contribution to the research.

%\author{A.B. Siddique, M.H. Maqbool\authornotemark{*}, Kshitija Taywade, and Hassan Foroosh\authornotemark{*}}

%\affiliation{%
%  \institution{University of Kentucky, University of Central Florida\authornotemark{*}}
  %\city{Riverside}
  %\state{CA}
 % \country{USA}
%}
%\email{siddique@cs.uky.edu, hasanmaqbool@knights.ucf.edu, kshitija.taywade@uky.edu, hassan.foroosh@ucf.edu}

\author{A.B. Siddique}
\affiliation{%
  \institution{University of Kentucky}
  \city{Lexington}
  \state{Kentucky}
  \country{USA}
}
\email{siddique@cs.uky.edu}

\author{M.H. Maqbool}
\affiliation{%
  \institution{University of Central Florida}
  \city{Orlando}
  \state{Florida}
  \country{USA}
}
\email{hasanmaqbool@knights.ucf.edu}

\author{ Kshitija Taywade}
\affiliation{%
  \institution{University of Kentucky}
  \city{Lexington}
  \state{Kentucky}
  \country{USA}
}
\email{kshitija.taywade@uky.edu}

\author{Hassan Foroosh}
\affiliation{%
  \institution{University of Central Florida}
  \city{Orlando}
  \state{Florida}
  \country{USA}
}
\email{hassan.foroosh@ucf.edu}
%%
%% By default, the full list of authors will be used in the page
%% headers. Often, this list is too long, and will overlap
%% other information printed in the page headers. This command allows
%% the author to define a more concise list
%% of authors' names for this purpose.
%\renewcommand{\shortauthors}{Anonymous Author(s)}

%%
%% The abstract is a short summary of the work to be presented in the
%% article.
\begin{abstract}
Task-oriented dialog systems enable users to accomplish tasks using natural language.
State-of-the-art systems respond to users in the same way regardless of their personalities, although personalizing dialogues can lead to higher levels of adoption and better user experiences.
Building personalized dialog systems is an important, yet challenging endeavor and only a handful of works took on the challenge.
% FJ .. rely on (the)(x) supervised learning approach(es)(v) ..
Most existing works rely on supervised learning approaches and require laborious and expensive labeled training data for each user profile. 
Additionally, collecting and labeling data for each user profile is virtually impossible.
In this work, we propose a novel framework, {\ourmodel}, to personalize task-oriented dialog systems capable of adapting to a wide range of user profiles in an unsupervised fashion using 
a zero-shot generalizable reward function.
{\ourmodel} uses a pre-trained GPT-2 as a backbone model and works in three phases. 
Phase one performs task-specific training. 
Phase two kicks off unsupervised personalization by leveraging the proximal policy optimization algorithm that performs policy gradients guided by the zero-shot generalizable reward function.
Our novel reward function can quantify the quality of the generated responses even for \emph{unseen} profiles.
The optional final phase fine-tunes the personalized model using a few labeled training examples.
We conduct extensive experimental analysis using the \myspecial{personalized bAbI dialogue} benchmark for five tasks and up to $180$ diverse user profiles. 
The experimental results demonstrate that {\ourmodel}, even when it had access to \emph{zero} labeled examples, outperforms state-of-the-art supervised personalization models and achieves competitive performance on BLEU and ROUGE metrics when compared to a strong fully-supervised GPT-2 baseline.

\end{abstract}

%%
%% The code below is generated by the tool at http://dl.acm.org/ccs.cfm.
%% Please copy and paste the code instead of the example below.
%%
%%
%% Keywords. The author(s) should pick words that accurately describe
%% the work being presented. Separate the keywords with commas.

%%
%% This command processes the author and affiliation and title
%% information and builds the first part of the formatted document.
\maketitle
\section{Introduction}
\label{intro}

\begin{figure*}[t!]
  \centering
  \includegraphics[width=\linewidth]{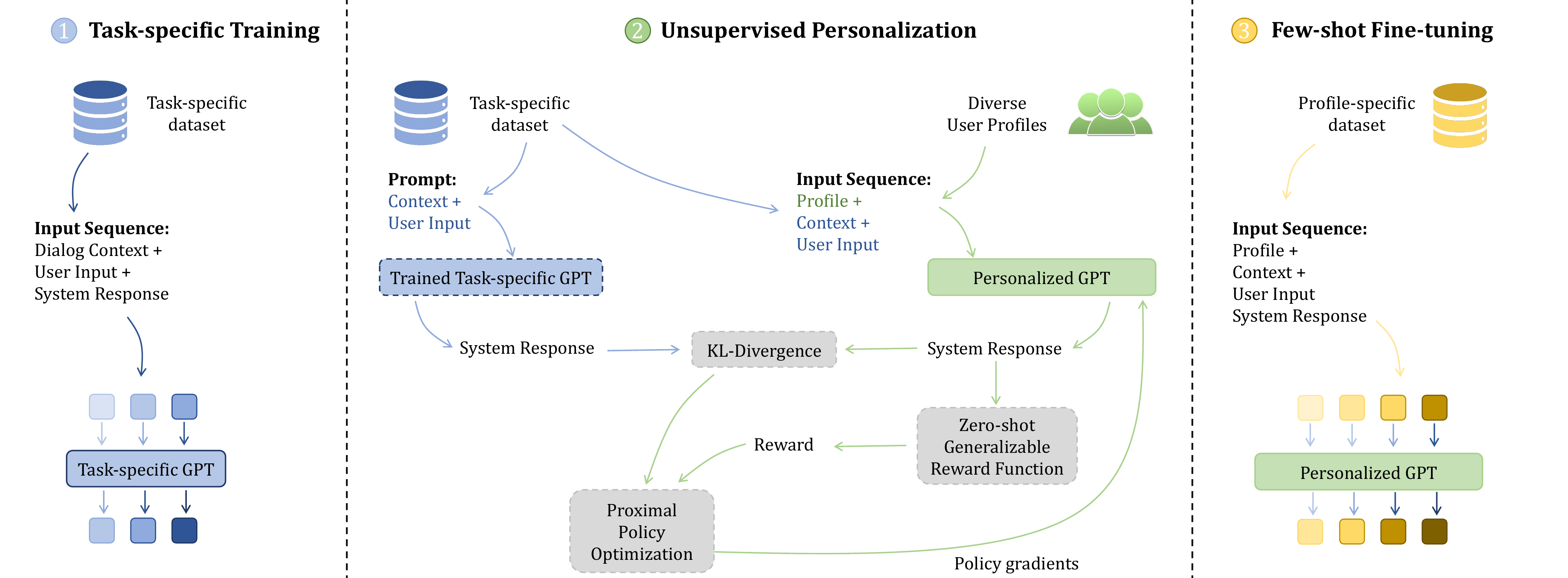}
  \caption{Overview of \ourmodel. The unsupervised personalization phase is at the core of the proposed framework.
  }
  \vspace{-5pt}
  \label{fig:overview}
\end{figure*}

Task-oriented dialog systems 
%(e.g.,~Amazon Alexa)
% FJ through --> using below
provide users with the ability to carry out tasks, such as reserving a table at a restaurant, using natural language~\cite{wen2017network}.
%Recently, researchers have increasingly focused on training end-to-end task-oriented dialog systems, as opposed to the pipeline approach~\cite{hosseini2020simple,bordes2016learning}.
% FJ Add references after '... pipeline approach' below
Contrary to the pipeline approach~\cite{chen2017survey}, researchers have increasingly focused on training end-to-end task-oriented dialog systems recently~\cite{hosseini2020simple,bordes2016learning}.
Such models generate responses exclusively based on the task-specific context of the dialog. Consequently, these models fail to adapt their responses to the diverse user personalities~\cite{joshi2017personalization}.
% FJ consider removing 'according' above
% FJ consider removing 'current' below, it's redundant
Specifically, state-of-the-art task-oriented dialog systems struggle to \myNum{i} adapt their conversation flows according to the active user’s personality, \myNum{ii} adjust their linguistic style, and \myNum{iii} handle ambiguities~\cite{herzig2017neural}.
In addition to presenting the choices to the user in an arbitrary or sequential order without taking the personality of the active user into account, task-oriented dialog systems use only task-specific, dull language.
It has been shown that adapting to the interlocutor improves communication efficiency~\cite{brown1965work,brown1987theory,kroger1992rules}.
Personalized task-oriented dialog systems can leverage profile information to expedite the interaction by understanding user’s actual information needs promptly, generate tailored responses by adapting linguistic variations, and properly address ambiguities by contextualizing nuanced queries -- a step towards delivering more human-like interactions~\cite{mo2018personalizing,yang2018investigating}. 
Personalizing task-oriented dialog systems without compromising the task completion accuracy is the focus of this work.

Earlier works use pre-training user profiles for intermediate supervision, as well as memory networks with copy mechanisms~\cite{qian2017assigning,herzig2017neural}.
The authors in~\cite{joshi2017personalization,luo2019learning} encode user information and conversation history using memory networks in an end-to-end fashion.
To synthesize personalized responses, \cite{zhang2019learning} utilized dynamic and static attention mechanisms in the end-to-end memory network.
For each user profile, these works require enormous amounts of labeled training data, which is time-consuming, expensive, and nearly impossible to acquire.
Recently, pre-trained language models have shown zero-shot capabilities in the natural language understanding and natural language generation tasks~\cite{du2021glam,brown2020language}, which suggests the possibility of developing personalized task-oriented dialog systems without
%the need for 
requiring labeled training data for each target user profile.
However, successfully exploiting the users' profiles and synthesizing personalized responses with no (or few) labeled training examples is a demanding task.

We introduce a novel framework for building \textbf{P}ersonalized \textbf{T}ask-\textbf{o}riented \textbf{D}ialog Systems, {\ourmodel}, that leverages the
%power of 
pre-trained language models (LMs), zero-shot (as well as few-shot) learning, and deep reinforcement learning.
% FJ Consider using {\ourmode} instead of 'the proposed framework' below and elsewhere
Guided by the proximal policy optimization (PPO) algorithm~\cite{schulman2017proximal,baselines} and a zero-shot generalizable reward function, the proposed framework
%FJ Consider '... and make them adapt to ...' instead of '... are capable of adapting to ...'
can personalize task-oriented dialog systems
%that are capable of adapting 
to diverse user profiles in an unsupervised fashion.
Figure~\ref{fig:overview} presents an overview of the framework that works in three phases and uses a pre-trained GPT-2~\cite{radford2019language} as a backbone model.
A task-specific training 
(e.g., reserving a table)
is performed in the first phase. Task-specific training datasets are generally available for a wide range of tasks in many domains~\cite{lee2021sgd,zang-etal-2020-multiwoz}, whereas personalized counterparts are 
%more or less 
practically
impossible to obtain. 
To overcome this challenge, we employ the unsupervised personalization phase. 
The deep reinforcement learning-based phase initializes a personalized GPT model from the task-specific GPT model (i.e., trained in phase one).
Then, it trains personalized GPT model based on \myNum{i} the appropriateness of the generated response for the given user profile, quantified by the zero-shot generalizable reward function; and \myNum{ii} fidelity of the response to the task, measured by the KL divergence between the responses generated by the task-specific and personalized models. Using the above signals, the PPO algorithm is employed to perform policy gradients.

We also propose a new reward function that allows quantifying the quality of the generated personalized responses not only for previously seen user profiles, but also for newly emerging unseen profiles. 
The zero-shot generalizable reward function uses pre-trained sentence transformers and contrastive representation learning to score the suitability of the response for the active user profile. To the best of our knowledge, this is the \emph{first work} that can  adapt the responses of task-oriented dialog systems to diverse user profiles in an unsupervised fashion.
To further improve the performance of the personalized task-oriented dialog systems, an \emph{optional} few-shot fine-tuning phase is introduced. This phase uses a few labeled training examples to adjust the responses for the given user profile, that can be employed or skipped depending on the availability of the labeled training data. 
Moreover, the number of shots can also be adjusted depending on the quantity of the available training examples.  

We perform thorough experimental evaluations on the 
only publicly available benchmark, 
\myspecial{personalized bAbI dialogue}
benchmark, for five tasks and up to 180 distinct user profiles in the restaurant domain.
The experimental results show that our proposed framework outperforms state-of-the-art supervised personalization models, even when given access to zero labeled training instances (i.e., few-shot fine-tuning phase is skipped). 
We also demonstrate that the proposed personalization approach achieves a competitive performance when compared to a strong supervised GPT-2 baseline model on the BLEU-4 and ROUGE-2 measures. 
% FJ Consider 'Furthermore, our human study confirms that {\ourmodel} is superior to existing supervised approaches.' instead of below.
Furthermore, the human study confirms the competitiveness of our unsupervised personalization framework to the other supervised approaches.

This work's contributions are summarized below:
\begin{itemize}[leftmargin=1.2\parindent,labelindent=-1pt, itemsep=-1pt]

    \item We propose an end-to-end framework for personalizing task-oriented dialog systems in an unsupervised way. To the best of our knowledge, this is the first work that has the unsupervised personalization capabilities.
    
    \item We introduce a zero-shot generalizable reward function that can guide the policy of the personalized task-oriented dialog systems to generate rich and personalized responses even for the unseen user profiles.
    
    \item We perform extensive experimental analysis using \myspecial{personalized bAbI dialogue} dataset and show that our framework consistently outperforms state-of-the-art supervised personalization models for up to $180$ unique user profiles on five tasks.
\end{itemize}
\section{Preliminaries}
\label{problem}
\subsection{Problem Formulation}
In a multi-turn task-oriented dialogue, 
$\bigU_\smallt$ is an input from the user and $\bigS_{\smallt}$ is a system's response 
at a turn \smallt. 
To generate a response $\bigS_{\smallt}$, all previous turns are concatenated to prepare dialog context $\bigC_\smallt = [\bigU_0, \bigS_0, \cdots,  \bigU_{\smallt-1}, \bigS_{\smallt-1}]$ and passed to the system as input along with the user’s current input $\bigU_\smallt$.
In a personalized task-oriented dialog system, at turn \smallt, the goal is to synthesize a response $\bigS^{\smalli}_\smallt$ adapted for a user profile $\bigP^\smalli \in \bigP = \{\bigP^0, \bigP^1, \cdots\}$.
The system’s response $\bigS^{\smalli}_\smallt$ is generated by conditioning on dialog context $\bigC_\smallt$, user’s current utterance $\bigU_\smallt$, and profile information $\bigP_\smalli$ for user $\smalli$, concatenated as a single sequence.
$$
\bigS^{\smalli}_\smallt = \ptod ([\bigP^\smalli ;  \bigC_\smallt ; \bigU_\smallt])
$$
In traditional (i.e., supervised) personalized task-oriented dialog systems, at turn \smallt, we are given $\smallm$ variants of the system response adapted for each user as: 
$ \{(\bigU_\smallt, \bigS^{\smalli}_\smallt)\}_{i=1}^\smallm$ for all $\smallm$ user profiles to train the models. 
The major disadvantage of such an approach is the unscalable requirement of having a large number of labeled training examples for each user profile; such data acquisition is expensive and time-consuming.
To overcome this challenge, we assume that, at turn $\smallt$, profile-specific response $\bigS^{\smalli}_\smallt $ $\forall \smalli$ is not available for model's supervision (i.e., unsupervised personalization). 
To allow for handling up to $\infty$ user profiles, we assume that the user profile is described via natural language text, in contrast to previous works that encode the features of the user profile via one-hot encoding and limits the model's expansion to new profile features. Naturally, describing user profiles using natural language takes care of the case where only partial information about a user profile is available.
Moreover, some tasks require interaction with knowledge base, we define the knowledge base tuples as $\bigK = [\smallk_1, \smallk_2, \cdots , \smallk_{\smalll}]$, where each tuple $\smallk_\smallb$ is defined using natural language and passed as additional input to the model where needed.

\subsection{Pre-trained Language Models}
\label{sec:lm}
% self-supervisedly
The Language models (e.g., GPT-2~\cite{radford2019language}, BERT~\cite{devlin2018bert}) are  trained utilizing massive amounts of text data in the unsupervised way. Since these models have millions of parameters, they have the capability to effectively capture both general semantic and syntactic information.
In this work, we utilize the pre-trained GPT-2 and MPNet~\cite{song2020mpnet} models.
We use GPT-2 as a base model, perform task-specific training, and then further train the model to synthesize personalized responses in an supervised way, guided by the novel reward function. 
The GPT-2 model has achieved state-of-the-art performance on many natural language generation benchmarks including conversation question answering~\cite{reddy2019coqa}, text summarization~\cite{nallapati2016abstractive}, and machine translation~\cite{lample2017unsupervised}, among others. 

We train a zero-shot generalizable reward function to score the acceptability of the generated responses for the given user profile using a contrastive loss function. 
The novel reward function uses 
%pre-trained sentence-Bert~\cite{reimers2019sentence} to acquire semantically accurate sentence embeddings. 
%We use the 
pre-trained MPNet~\cite{song2020mpnet} as a basic building block to acquire semantically accurate embeddings.
%in sentence-Bert.
%The sentence-Bert uses BERT~\cite{devlin2018bert} as a basic building block.
%The sentence-Bert model has produced cutting-edge results on semantic textual
%similarity benchmarks, such as sentiment prediction~\cite{pang2005seeing,socher2013recursive}, opinion polarity~\cite{wiebe2005annotating}, and paraphrase detection~\cite{dolan2004unsupervised}.
The MPNet model has produced cutting-edge results on several natural language processing tasks including GLUE~\cite{wang2018glue}, SQuAD~\cite{rajpurkar2016squad,rajpurkar2018know}, RACE~\cite{lai2017race}, and sentiment prediction~\cite{maas2011learning} benchmarks.
%
%In the following, we provide a brief overview of the causal language modeling (i.e., employed by GPT-2), masked language modeling, and next sentence prediction (i.e., used by BERT).
%
In the following, we provide a brief overview of the GPT-2 and MPNet models.

%\stitle{Causal Language Modeling.} 
\stitle{GPT-2.} 
The GPT-2 model is pre-trained for autoregressive generation (i.e., predicting the next word) on the WebText dataset (i.e., 40 GB of text) and adapts a transformer-based neural architecture~\cite{vaswani2017attention}.
Suppose we have a natural language sequence  $(\smalls_1, \cdots , \smalls_\smalln)$ where symbol $\smalls_\smalli$ is drawn from a fixed set of symbols. The sequential ordering of language leads to factorizing the joint probabilities over symbols as a product of conditional probabilities~\cite{bengio2003neural}, as given below.
$$
p(\smalls) = \prod_{i=1}^\smalln p(\smalls_\smalli | \smalls_1, \cdots ,\smalls_{\smalli-1})
$$
Using this approach, it is possible to estimate $p(\smalls)$ and any conditionals of the form 
$p(\smalls_{\smalli - k}, \cdots , \smalls_\smalli | \smalls_1, \cdots , \smalls_{\smalli - k - 1} )$, and perform tractable sampling. 
%Accordingly, the GPT-2 model is trained with parameters $\theta$ to minimize the negative log-likelihood over the entire dataset:
%$$
%- \sum_{i=1}^{n} \log {p_{\theta}(\smalls_{\smalli} | \smalls_1, \cdots ,\smalls_{\smalli-1})} 
%$$
  
\stitle{MPNet.} 
BERT does not account for interdependence among predicted tokens, whereas complete position information is not used by XLNet~\cite{yang2019xlnet}, though dependency among tokens is considered. 
The MPNet model exploits the benefits of masked language modeling (MLM) (i.e., employed by BERT) and permuted language modeling (PLM) (i.e., used by XLNet) and eliminates their shortcomings.
It brings out the best of both worlds: by using PLM, it exploits the predicted token's dependencies, and, at the same time, uses the full position information of a sentence from MLM to enable a full view of the sentence.
It has been pre-trained on BooksCorpus~\cite{zhu2015aligning}, OpenWebText, CC-News, Stories~\cite{trinh2018simple}, and Wikipedia (i.e., over 160GB data).
For a given sequence $(\smalls_1, \cdots, \smalls_\smalln)$, where permutations of set $\{1, \cdots, \smalln\}$ is represented by $\mathcal{Z}_\smalln$, 
the $\smallt$-th element of $z$ by $z_\smallt$,
the first $t-1$ element of $z$ by $z_{<t}$,
the number of non-predicted tokens by $c$, and
the mask tokens $\rm{[M]}$ in position $z_{>c}$ by $M_{z_{>c}}$.
The MPNet is trained for the following objective:
$$
\mathbb{E}_{z \in \mathcal{Z}_n} \sum\limits_{t=c+1}^{n} \log p(\smalls_{z_\smallt}|\smalls_{z_{<\smallt}}, M_{z_{>c}};\theta)
$$

\subsection{Reinforcement Learning Paradigm}
The reinforcement learning paradigm has been extensively studied for unsupervised learning.
Methods that use policy gradients compute an estimator of the gradient and then plug it into a stochastic gradient ascent algorithm.
It is common to optimize the policy $\pi$ by maximizing the expected reward $r \in \realR$ for the generated sequence $\bigY = (y_1, \cdots, y_n)$ with length $n$, given the input sequence $\bigX = (x_1, \cdots, x_m) $ with length $m$, that is sampled from data distribution $\bigD$. We can optimize the expected reward as follows:
$$
\mathbb{E}_{\pi}[\smallr] = \mathbb{E}_{x \sim \bigD, y \sim \pi(\cdot | x)} [r(x, y)]
$$

The PPO algorithm introduced clipped surrogate objective, in addition to, the penalty on the KL divergence.
The objective function is modified using the KL divergence penalty, instead of making it a hard constraint like in the trust region policy optimization algorithms~\cite{schulman2015trust}. The PPO updates its policy, at step $k$ via:
$$
\theta_{k+1} = \arg\max_\theta \mathbb{E}_{s,a \sim \pi_{\theta_k}} [ \mathcal{L} (s, a, \theta_k, \theta)]
$$ 
where $s$ and $a$ represent the state and action, respectively.
In this work, we employ PPO algorithm~\cite{baselines} to perform policy gradients, that has been shown to be scalable (e.g., for large language models), data-efficient, and robust (i.e., without excessive hyperparameter tuning)~\cite{bellemare2013arcade}.

\begin{figure*}[t!]
  \centering
  \includegraphics[width=0.96\linewidth]{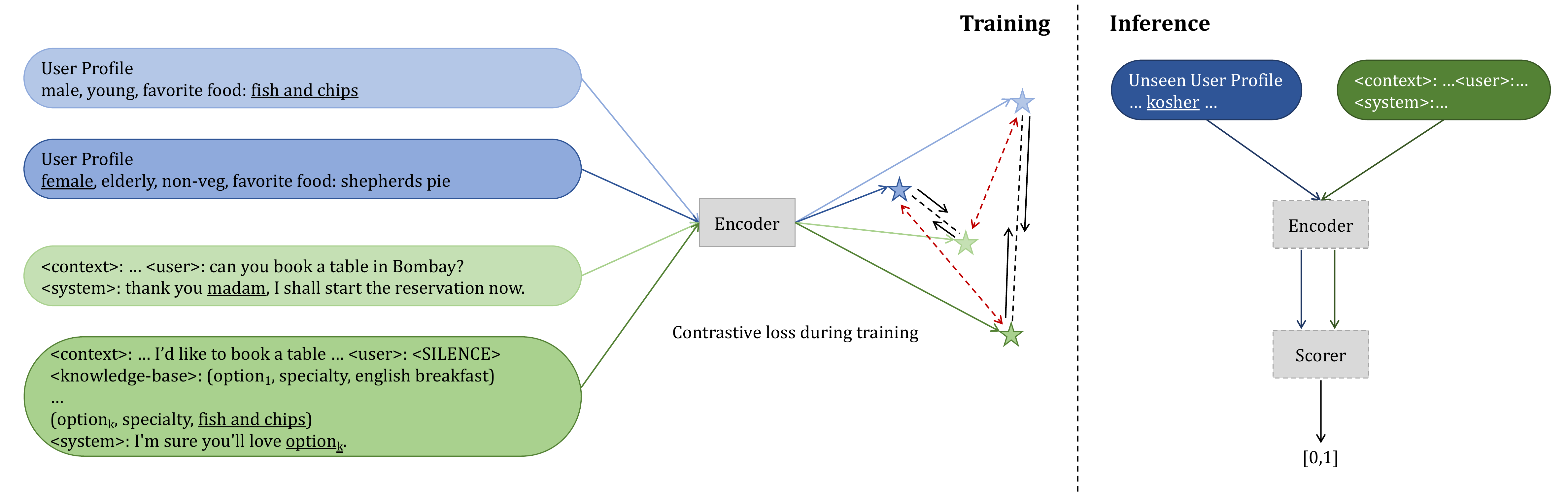}
  \caption{Overview of the training and inference process for the zero-shot generalizable reward function.
  }
  \vspace{-10pt}
  \label{fig:zs}
\end{figure*}

\section{Personalization Framework: \ourmodel}
\label{model}
This work presents a new framework for developing personalized dialog systems that works in three phases. 
A pre-trained GPT-2 model serves as the backbone model for the framework.
In the first phase, the base GPT-2 model is optimized via task-specific training.  
The phase two, referred to as unsupervised personalization phase, employs deep reinforcement learning to adapt the system responses to a wide range of user profiles guided by the zero-shot generalizable reward function (i.e., presented in Figure~\ref{fig:zs}) and the trained task-specific GPT model. 
The \emph{optional} phase three fine-tunes the personalized GPT model using a few supervised training examples to further improve the performance. Figure~\ref{fig:overview} summarizes the proposed unsupervised personalization framework.

\subsection{Phase One: Task-specific Training}
\label{sec:phaseone}

\begin{figure}[t!]
  \centering
  \includegraphics[width=0.96\linewidth]{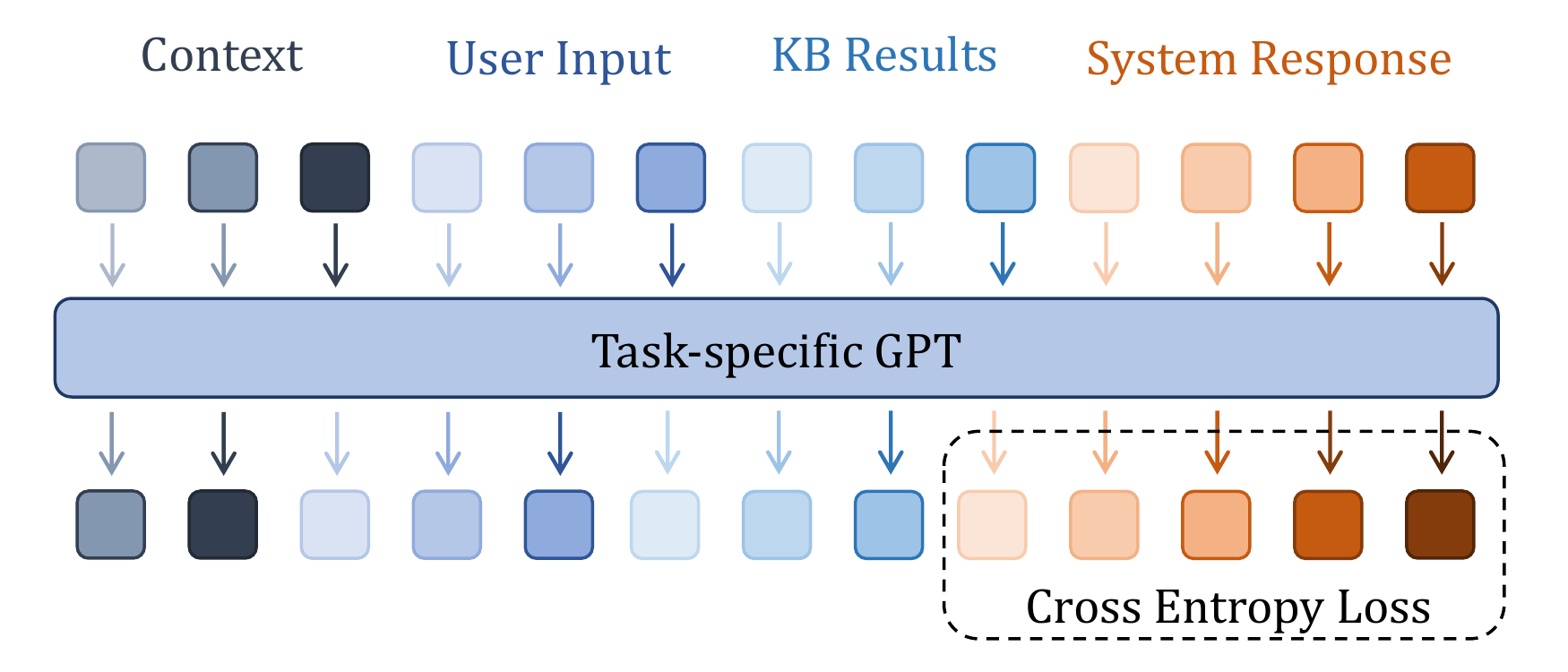}
  \caption{The task-specific training of the GPT-2 model.
  }
  \vspace{-12pt}
  \label{fig:task}
\end{figure}

We leverage the power of the pre-trained language models by intializing the phase one of our framework with a pre-trained GPT-2 model. The details of the pre-trained model are as follows. 
The model~\cite{radford2019language} was pre-trained on the WebText dataset and has 774 million parameters. 
Using byte pair encoding, the vocabulary size is 50,257 tokens; capitalization and punctuation were preserved~\cite{sennrich2015neural}. 
The model is built on the transformer's decoder stack~\cite{vaswani2017attention}, and it has 36 layers, 20 heads, and an embedding size of 1280.
The task-specific training of the model is performed using causal language modeling (see Section ~\ref{sec:lm} for details). 
Figure~\ref{fig:task} presents the task-specific training of the model.
Given a dialog context $\bigC_\smallt$, user's current utterance $\bigU_\smallt$, and (optional) knowledge base search result tuples $\bigK$ at turn $\smallt$, the probability of system's response $\bigS_\smallt$ with length $\smalln$ can be defined as:
$$
p(\bigS_\smallt | \bigC_\smallt, \bigU_\smallt, \bigK) = \prod_{i=1}^\smalln p(\smalls_\smalli | \smalls_{<\smalli}, \bigC_\smallt, \bigU_\smallt, \bigK)
$$

We train the model by calculating the cross-entropy loss by maximizing the log-likelihood of the system response conditioned on the dialog context, user's input, and knowledge base tuples. If the task does not require interaction with the knowledge base, the search query is not performed nor the generation is conditioned on the resultant tuples.
%from the knowledge base. 
The output of phase one is the trained task-specific GPT model. %It is important to highlight that this trained model cannot (automatically) adapt to different user profiles,  in its current state.  

\subsection{Phase Two: Unsupervised Personalization}
\label{sec:phasetwo}
This phase initializes the personalized GPT model with the trained task-specific GPT model (i.e., output of phase one). 
The personalized GPT model is trained for personalization in the unsupervised way. 
The two critical training signals are provided by \myNum{i} the zero-shot generalizable reward function that quantifies whether the output of the personalized model is apprpriate for the given user profile; and \myNum{ii} the KL divergence between the personalized and task-specific model's distributions to ensure that the output of the personalized model does not deviate too much from the task-specific model (i.e., it still accomplishes the task with high accuracy). 

In the following, we describe the details of the novel reward function and KL divergence. Then, we detail the training process for the unsupervised personalization phase.

\begin{figure*}[t!]
  \centering
  \includegraphics[width=0.96\linewidth]{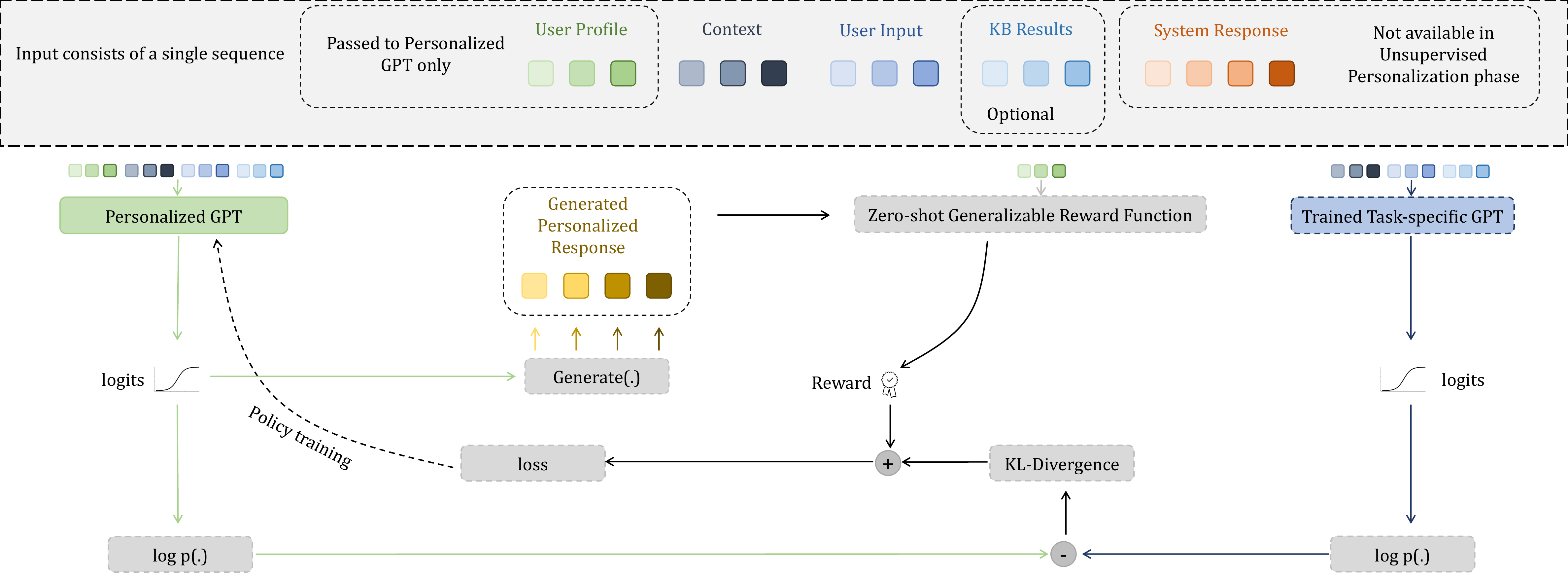}
  \caption{Phase two of the framework: Unsupervised Personalization.
  }
  \label{fig:personalization}
  \vspace{-5pt}
\end{figure*}

\stitle{Zero-shot Generalizable Reward Function.}
The zero-shot generalization is enabled by the unsupervised representations provided by the powerful pre-trained language model MPNet and the contrastive loss function~\cite{hadsell2006dimensionality}. 
The training and inference process of the reward function is shown in Figure~\ref{fig:zs}.
At a dialog turn $\smallt$, we concatenate the dialog context $\bigC_\smallt$, user's current input $\bigU_\smallt$, the (optional) knowledge base search result tuples $\bigK$, and the system's response $\bigS^{\smalli}_\smallt$ for the user $\smalli$ and acquire their representation $\bigH^{\smalli}_\smallt$. Similarly, we encode the user profile information $\bigP_\smallj$ for the user $\smallj$ to get a corresponding representation $\bigU^\smallj$.
If a pair of encodings had a positive corresponding label (i.e., the system response is appropriate for the given user profile), then the contrastive loss function would reduce their distance, and if a negative label were given, it would increase their distance. We generate positive training examples by setting $\smalli == \smallj$ and negative examples are generated by setting $\smalli \neq \smallj$. The training loss can be defined as:
$$
\mathcal{L}_{\smalli,\smallj}
  =-\log{
  \frac{\text{exp}\left(\bigH^{\smalli}_\smallt \bigcdot \bigU^\smallj /\tau\right)}{\sum\limits_{\smallq \in \bigQ }\text{exp}\left(\bigH^{\smalli}_\smallt \bigcdot  \bigU^\smallq / \tau\right)}
  }
$$
where the $\bigcdot$ represents the scoring function, $\tau\in\mathcal{R}^+$ is a scalar parameter for temperature, and $\bigQ$ is the set of negative pairs, i.e., $\smalli \neq \smallj$.
To train a classifier that works in the zero-shot setting, we select a subset of user profiles (i.e., \emph{seen} profiles) and use them to train the classifier. 
The pre-trained MPNet has the capability to generate rich, accurate, and high-quality embeddings even for the \emph{unseen} user profiles or unseen knowledge base entries, since both the user profile and knowledge base tuples are described using natural language. 
For example, the model can produce precise embeddings for an unseen user profile who prefers ``kosher'' food, because it has already learned the contextual usage of a large number of words (e.g., MPNet has a vocabulary size of 30,527) in the pre-training process.
The scoring function learns to score close to one, the matching pairs (i.e., the system response is appropriate for the given profile), and zero otherwise.

Our zero-shot generalizable reward function follows the Sentence-BERT~\cite{reimers2019sentence} that employs siamese and triplet network structures~\cite{schroff2015facenet}, leverages contrastive loss, and dot product is used as the scoring function. 
To generate input encoding, we use the pre-trained \myspecial{all-mpnet-base-v2} that has been trained on over one billion training pairs and produces 768 dimensional normalized embeddings for the input by \myspecial{mean pooling}.
For every positive training pair, two negative training examples are generated.
At inference time, the trained zero-shot generalizable reward function provides a scalar reward, $\smallr \in $~[0,1] that quantifies the suitability of the system's responses for both previously seen and newly emerging unseen user profiles.

\stitle{KL Divergence.}
To ensure that the personalized policy does not diverge too much from the trained task-specific model, we use an additional reward signal by calculating the KL divergence between the personalized policy and the task-specific policy (i.e., the model trained in phase one).
That is, keeping close to the task-specific model is rewarded, whereas big KL divergences are penalized.
We denote the distributions of the task-specific and personalized models by $p_{\one}$ and $p_{\two}$, respectively. At dialog turn $\smallt$, the KL divergence can be calculated as:
$$
KL =
\mathbb{E}_{\bigS^\smalli_\smallt \sim p_{\two}} [
\log
p_{\two}(\bigS^\smalli_\smallt | \bigP^\smalli, \bigC_\smallt, \bigU_\smallt, \bigK)
- \log
p_{\one}(\bigS_\smallt | \bigC_\smallt, \bigU_\smallt, \bigK)
]
$$
where $\bigS_\smallt$ is the task-specific response and $\bigS^\smalli_\smallt$ is the system's response adapted for the user $\smalli$. The final $reward$ can be combined as given below:
$$
reward = \smallr + \beta \times KL
$$
where $\beta \in [0, -1]$ is the penalty coefficient and decides the weight of the KL divergence. We use
adaptive KL Penalty coefficient and
initialize $\beta = -0.2 $ in our experiments .

\stitle{Training Details.}
To start with the unsupervised personalization phase, we initialize our personalized model $p_\two$ = $p_\one$ and then adapt $p_\two$ to synthesize the personalized responses for a wide range of user profiles using deep reinforcement learning.
The personalized model is fine-tuned via PPO algortihm from~\cite{baselines} with the final $reward$ (i.e., a combination of KL divergence and a score from zero-shot generalizable reward function). 
The expected reward for a response $\bigS^\smalli_\smallt$ for the user $\smalli$ at a dialog turn $\smallt$ can be written as:
$$
\mathbb{E}_{p_\two}[reward] = \mathbb{E}_{\bigU_\smallt \sim \omega, \bigS^\smalli_\smallt \sim p_\two(\cdot | \bigP^\smalli, \bigC_\smallt, \bigU_\smallt, \bigK)} [reward(\bigP^\smalli, \bigS^\smalli_\smallt)]
$$
where $\omega$ represents a given task, the model $p_\two$ is being trained for. The personalized model is trained for up to 600,000 episodes using Adam optimizer~\cite{kingma2014adam} with a learning rate of $1.41 \times 10^{-5}$.

The output of this phase is a personalized model that can generate responses that are not only specific to the task, but are also adapted for the given user profile. It is important to recall that the unsupervised personalization phase does not use any personalized variants of the responses for training the model. It is exclusively trained in the unsupervised setting, guided by the zero-shot generalizable reward function and KL divergence between the distributions of the task-specific and personalized models.

\begin{table*}[t!]
\centering
\caption{Datasets statistics.}
\label{tab:dataset}
\begin{tabular}{llccccc}
\toprule
\multicolumn{2}{l}{\textbf{Dataset} }           & \multicolumn{1}{c}{\textbf{Task 1}} & \multicolumn{1}{c}{\textbf{Task 2}} & \multicolumn{1}{c}{\textbf{Task 3}}  & \multicolumn{1}{c}{\textbf{Task 4}} & \multicolumn{1}{c}{\textbf{Task 5} } \\ \hline
\multirow{2}{*}{bAbI dialogue}              & Number of dialogs         & \multicolumn{1}{c}{4000}                                 & \multicolumn{1}{c}{4000}                                 & \multicolumn{1}{c}{4000}                                 & \multicolumn{1}{c}{4000}                                 & \multicolumn{1}{c}{4000}                                 \\
%                                           & \# Training Dialogs & \multicolumn{1}{c}{1000}                                 & \multicolumn{1}{c}{1000}                                 & \multicolumn{1}{c}{1000}                                 & \multicolumn{1}{c}{1000}                                 & \multicolumn{1}{c}{1000}                                 \\
%                                           & \# Validation Dialogs & \multicolumn{1}{c}{1000}                                 & \multicolumn{1}{c}{1000}                                 & \multicolumn{1}{c}{1000}                                 & \multicolumn{1}{c}{1000}                                 & \multicolumn{1}{c}{1000}                                 \\
%                                           & \# Test Dialogs & \multicolumn{1}{c}{1000}                                 & \multicolumn{1}{c}{1000}                                 & \multicolumn{1}{c}{1000}                                 & \multicolumn{1}{c}{1000}                                 & \multicolumn{1}{c}{1000}                                 \\
%                                           & \# Test-OOV Dialogs & \multicolumn{1}{c}{1000}                                 & \multicolumn{1}{c}{1000}                                 & \multicolumn{1}{c}{1000}                                 & \multicolumn{1}{c}{1000}                                 & \multicolumn{1}{c}{1000}                                 \\
                                           & Avg. dialog turns & \multicolumn{1}{c}{6.0}                                 & \multicolumn{1}{c}{9.5}                                 & \multicolumn{1}{c}{9.9}                                 & \multicolumn{1}{c}{3.5}                                 & \multicolumn{1}{c}{18.4}                                 \\ \hline
\multirow{4}{*}{\begin{tabular}[c]{@{}l@{}}Personalized\\  bAbI dialogue\end{tabular}} & Number of dialogs         &  24000                                                    &    24000                                                  &    48000                                                  &     24000                                                 &    48000                                                  \\
%& \# Training Dialogs & \multicolumn{1}{c}{6000}                                 & \multicolumn{1}{c}{6000}                                 & \multicolumn{1}{c}{12000}                                 & \multicolumn{1}{c}{6000}                                 & \multicolumn{1}{c}{12000}  \\
%& \# Validation Dialogs & \multicolumn{1}{c}{6000}                                 & \multicolumn{1}{c}{6000}                                 & \multicolumn{1}{c}{12000}                                 & \multicolumn{1}{c}{6000}                                 & \multicolumn{1}{c}{12000}  \\
%& \# Test Dialogs & \multicolumn{1}{c}{6000}                                 & \multicolumn{1}{c}{6000}                                 & \multicolumn{1}{c}{12000}                                 & \multicolumn{1}{c}{6000}                                 & \multicolumn{1}{c}{12000}  \\
%& \# Test-OOV Dialogs & \multicolumn{1}{c}{6000}                                 & \multicolumn{1}{c}{6000}                                 & \multicolumn{1}{c}{12000}                                 & \multicolumn{1}{c}{6000}                                 & \multicolumn{1}{c}{12000}  \\
                                            & Avg. dialog turns &   6.0                                                   &   9.5                                                   &  11.8                                                    &     3.5                                                 &   20.3                                                   \\
                                            & Number of user profiles        &6                                                      &6                                                      &    180                                                  &   6                                                   & 180 \\
                                            & 
                                            Avg. dialogs per profile        &   4000                                                   &   4000                                                   &   267                                                   &    4000                                                  &  267                         \\ \bottomrule
\end{tabular}
\vspace{-9pt}
\end{table*}

\subsection{Phase Three: Few-shot Fine-tuning}
The optional phase three uses a few labeled training examples to calibrate the personalized model (i.e., trained in phase two in the unsupervised setting) for the given user profile in the supervised setting. 
The probability for system's response $\bigS^\smallj_\smallt$ with length $\smalln$, for a given user $\smallj$, at dialog turn $\smallt$ can be defined as:
$$
p(\bigS^\smallj_\smallt | \bigP^\smallj, \bigC_\smallt, \bigU_\smallt, \bigK) = \prod_{i=1}^\smalln p(\smalls_\smalli | \smalls_{<\smalli}, \bigP^\smallj, \bigC_\smallt, \bigU_\smallt, \bigK)
$$

We call this phase \emph{optional}, since it can be employed or skipped based on the availability of the labeled variants for the given user profile.
Moreover, the number of shots can also be adjusted depending on the quantity of the available training examples. In our experiments, we present results with the following number of shots: 0 (i.e., we skip this phase), 1, 5, 10, and 20.

\section{Experimental Setup}
\label{experiments}

In this section, we describe the task-specific and personalization datasets, methodology of evaluation, competing methods, and the implementation details of our framework {\ourmodel}.

\subsection{Datasets}
\label{datasets}
We used one task-specific task dataset \myspecial{bAbI dialogue}~\cite{bordes2016learning} that trains our model in phase one. The personalized counterpart, called \myspecial{personalized bAbI dialogue}~\cite{joshi2017personalization}, is used to train all the supervised competing models. Our proposed framework adapts to diverse user profiles in the unsupervised setting.
To the best of our knowledge, \myspecial{personalized bAbI dialogue} is the \emph{only} publicly available personalization benchmark for task-oriented dialog systems. Table~\ref{tab:dataset} presents important statistics for both datasets. Both datasets are in the restaurant domain and consist of five tasks.

\stitle{Task 1: Issue API calls.}
This task involves extracting values of all the required slots (a.k.a. values for query parameters, e.g., \myspecial{cuisine} = \myspecial{spanish}) from natural language utterances and successfully making an API call. In this task, the personalization involves understanding and adapting the linguistic variations for a given user profile (e.g., male vs female).

\stitle{Task 2: Update API calls.}
This task includes updating the values for certain slots, if the user wishes to do so. For example, a user's request in natural language, ``Instead could it be in a cheap price range in Madrid?'', should update the current API call: \myspecial{api\_call(cuisine=french, city=paris, party\_size=four, price\_range=expensive)} to the call: \myspecial{api\_call(cuisine=french, city=madrid, party\_size=four, price\_range=cheap)}. Similarly to task one, personalization task two mainly deals with the style adaptations.

\stitle{Task 3:  Display Options.}
This task requires displaying relevant options from the knowledge base using the search results from API call. The personalization task involves adapting certain linguistic style as well as understanding user's taste and restaurant's specialities, among others, and making appropriate suggestions based on the active user's profile. Unsupervised personalization for this task is the most challenging part of this work. 

\stitle{Task 4: Provide extra information.}
The user's acceptance of an option entails asking for extra information (e.g., \myspecial{phone\_number}) from the system. The personalization for task four calls for resolving ambiguities efficiently along with the style adaptation. For example, asking for contact information could refer to \myspecial{phone\_number} or \myspecial{social\_media} depending on the active user (e.g., elederly vs young).

\stitle{Task 5: Conduct  Full dialogs.}
This task is about conducting the full dialogue that covers tasks 1-4 successfully. Similarly, personalization task includes, but not limited to: \myNum{i} adjusting the conversation flow to the active user’s personality, \myNum{ii} adapting the linguistic style, and \myNum{iii} dealing with nuances effectively. 

The \myspecial{personalized bAbI dialogue} dataset contains two test sets: a standard test set and a test set - \myspecial{OOV} (Out Of Vocabulary). We conduct extensive experiments on both test sets for all the five tasks for up to 180 diverse user profiles.

\begin{table*}[t!]
\caption{F1 scores for task completion.}
\label{tab:f1}
\centering
\begin{tabular}{llccccc}
\toprule
\textbf{Approach}                           & \textbf{Models} & \multicolumn{1}{l}{\textbf{Task 1}} & \multicolumn{1}{l}{\textbf{Task 2}} & \multicolumn{1}{l}{\textbf{Task 3}} & \multicolumn{1}{l}{\textbf{Task 4}} & \multicolumn{1}{l}{\textbf{Task 5}} \\ \hline
\multirow{9}{*}{Supervised} & MemNN-org       & 99.63                               & 99.81                               & 98.87                               & 98.87                               & 85.10                               \\
                                            & MemNN-split     & 85.66                               & 85.83                               & 84.89                               & 84.89                               & 87.28                               \\
                                            & PMemN2N         & 99.70                               & 99.93                               & 98.91                               & 98.97                               & 95.33                               \\
                                            & Mem2Seq-org     & 99.68                               & 99.68                               & 98.28                               & 99.68                               & 80.41                               \\
                                            & Mem2Seq-split   & 99.62                               & 99.62                               & 98.52                               & 99.62                               & 82.19                               \\
                                            & Mem2Seq-att     & 99.66                               & 99.66                               & 98.46                               & 99.66                               & 82.38                               \\
                                            & GLMP            & 99.45                               & 99.45                               & 98.48                               & 99.45                               & 86.20                               \\
                                            & CoMemNN         & 99.65                               & 99.65                               & 98.61                               & 99.65                               & 98.13                               \\
                                            & Supervised-GPT  & \underline{99.72}                      & \textbf{99.96}                      & \underline{99.02}                      & \textbf{99.96}                      & \textbf{98.21}                      \\
                                            \hline
Unsupervised Personalization                & PToD-0   (This work)       & 99.69                               & 99.86                               & 98.92                               & 99.88                               & 98.14                 \\ \hline
\multirow{2}{*}{Few-shot Personalization} & Few-shot GPT       & 98.12                               & 99.08                               & 97.71                               & 97.32                               & 91.23                               \\
                    & {\ourmodel}  (This work)          & \textbf{99.74}                               & \underline{99.94}                               & \textbf{99.03}                               & \underline{99.94}                               & \underline{98.17}   \\
\bottomrule
\end{tabular}
\vspace{-8pt}
\end{table*}

\subsection{Evaluation Methodology}
\label{testing}

To demonstrate the effectiveness of {\ourmodel}, we evaluate our framework and all the competing methods for \myNum{i}~task completion and \myNum{ii}~personalization of the dialog for the given user profile.

\stitle{Task Completion.}
To quantify the performance for the task completion, we compute the F1 scores and present evaluation results for all the models for all five tasks.

\stitle{Personalization.}
The main task for the proposed framework is to personalize the task-oriented dialog systems in the unsupervised way. To evaluate the efficacy of the framework and how it compares to the other supervised approaches, we use BLEU-4 and ROUGE-2 scores. The BLEU~\cite{papineni2002bleu} and ROUGE~\cite{hovy2006automated} metrics have been extensively used for natural language generation tasks.
Human judgment and BLEU scores show a very strong correlation. 
The BLEU-n (n $\in \{1,2,3,4\}$) score $\in [0,100]$ measures the proportion of n-grams in the generation that also occurs in the reference.
ROUGE, on the other hand, is a recall-based measure that quantifies n-gram overlap between the generation and the reference.
Moreover, we also conduct a user study on a randomly selected 300 responses generated by the top performing supervised models and our proposed unsupervised personalization framework.

\subsection{Competing Methods}
\label{baselines}
We compare against the following state-of-the-art (SOTA) personalization models and GPT-2-based strong baselines:
\begin{description}[leftmargin=1.2\parindent,labelindent=-3.5pt, itemsep=-1pt]
\item \textbf{MemNN~\cite{joshi2017personalization}:}
The response selection-based approach proposes to use the memory network to encode dialog content and user profile information using a concatenation of the profile information and dialog memory (i.e., MemNN-org) and using split memory for the profile information and concatenating hidden states (i.e., MemNN-split).

\item \textbf{PMemN2N~\cite{luo2019learning}:}
The memory network-based method facilitates the model's personalization by combining the style information of the user attributes in the encoder.

\item \textbf{Mem2Seq~\cite{madotto2018mem2seq}:}
An end-to-end approach that proposes to use memory network in the encoder and employs RNN-based decoder for query generation and memory network for personalized response generation. This work proposes three variants of the models, called Mem2Seq-org, Mem2Seq-split, and Mem2Seqatt.

\item \textbf{GLMP~\cite{wu2019global}:}
Based on Mem2Seq, this model includes local and global encoders to share external knowledge efficiently.

\item \textbf{CoMemNN~\cite{pei2021cooperative}:}
This work proposes cooperative memory network and assumes that only partial user profile information is available. This approach does not generate response, instead relies on the response selection. In our experiments, we provided the model with 100\% user profile information for a fair comparison.

\item \textbf{Supervised GPT:}
Since none of the SOTA personalized models follow SOTA transformers architecture, we also trained a supervised GPT-2 model. This model was trained in the same fashion as our phase three except it was trained on all the training examples of the dataset, thus serves as a strong supervised baseline.
\item \textbf{Few-shot GPT:}
Due to the unavailability of any unsupervised approach for comparison and coming up with a reward function is non-trivial, we also trained a few-shot GPT-2 model. This model follows same training process, except phase two (i.e., unsupervised personalization) is skipped to demonstrate the effectiveness of the phase two of the proposed framework.

\end{description}

\subsection{Implementation Details}
\label{implementation}
We use the pre-trained GPT-2 model as a backbone model that is trained in all the three phases of the framework. 
The phase one trains the task-specific model for 3 epochs using cross-entropy loss and Adam optimizer, with a batch size of 8, and a learning rate of $5 \times 10^{-5}$. Other parameters are as follows:
\myspecial{warmup\_steps=100}, \myspecial{weight\_decay=0.01},
\myspecial{max\_length=1024}. 
The zero-shot generalizable reward function uses a pre-trained MPNet for input encoding.
It is trained for 3 epochs using contrastive loss on 50\% of the user profiles on every task and the remaining 50\% profiles are considered unseen. 
The phase two uses the same parameters as phase one, except batch size of 4 was used because of the GPU memory limitations (and a learning rate of $1.41 \times 10^{-5}$). Similarly, phase three uses same parameters, except a smaller learning rate of  $5 \times 10^{-7}$ was used and up to 20 training examples were made available for training.
We present two variants of our model: \myNum{i} PToD-0 does not use phase three (i.e., personalized model is only trained in the unsupervised setting) and \myNum{ii} {\ourmodel} that uses 20 training examples in the phase three.

\section{Results}
\label{results}

\begin{table*}[t!]
\caption{BLEU scores and ROUGE scores for personalization for all five tasks.}
\label{tab:bleu}
\centering
\begin{tabular}{llcc|cc|cc|cc|cc}
\toprule
\multirow{2}{*}{\textbf{Approach}} & \multirow{2}{*}{\textbf{Models}} & \multicolumn{2}{c|}{\textbf{Task 1}} & \multicolumn{2}{c|}{\textbf{Task 2}} & \multicolumn{2}{c|}{\textbf{Task 3}} & \multicolumn{2}{c|}{\textbf{Task 4}} & \multicolumn{2}{c}{\textbf{Task 5}} \\ \cline{3-12} 
                                   &                                  & BLEU           & ROUGE           & BLEU           & ROUGE           & BLEU           & ROUGE           & BLEU           & ROUGE           & BLEU           & ROUGE          \\ \hline
\multirow{6}{*}{Supervised}          & Mem2Seq-org                      & 60.12            & 64.82             & 65.54            & 69.83             & 57.74            & 62.73             & 59.07            & 63.32             & 64.23            & 59.39            \\
                                   & Mem2Seq-split                    & 60.30            & 63.82             & 64.92            & 68.60             & 58.07            & 62.43             & 59.20            & 63.03             & 64.11            & 58.73            \\
                                   & Mem2Seq-att                      & 62.26            & 71.17             & 67.15            & 75.84             & 59.84            & 69.59             & 61.29            & 69.74             & 66.02            & 66.17            \\
                                   & GLMP                             & 61.25            & 70.81             & 66.40            & 75.46             & 59.07            & 68.93             & 59.66            & 70.13             & 64.91            & 65.74            \\
                                   & CoMemNN                          & 68.67            & 77.71             & 73.83            & 82.67             & 65.77            & 75.72             & 67.58            & 76.85             & 72.23            & 72.53            \\
                                   & Supervised-GPT                   & \textbf{75.71}            & \underline{78.42}             & \textbf{80.61}            & \textbf{83.38}             & \underline{73.21}            & \textbf{76.46}             & \textbf{74.64}            & \underline{77.11}             & \textbf{80.01}            & \underline{73.61}            \\ \hline
Unsupervised           & PToD-0  (This work)                         & 70.84            & 75.02             & 75.75            & 79.85             & 68.44            & 72.93             & 69.72            & 73.69             & 75.12            & 70.21            \\
\hline
\multirow{2}{*}{Few-shot}          & Few-shot GPT                      & 40.21            & 46.71             & 33.17            & 39.32             & 27.17            & 22.78             & 39.20            & 33.25             & 24.12            & 29.31            \\
       & {\ourmodel}   (This work)                          & \underline{75.64}            & \textbf{78.46}             & \underline{80.55}            & \underline{83.29}             & \textbf{73.24}            & \underline{76.37}             & \underline{74.52}            & \textbf{77.13}             & \underline{79.92}            & \textbf{73.65}    \\
\bottomrule
\end{tabular}
\vspace{-5pt}
\end{table*}

In this section, we present quantitative as well as qualitative analysis. We first present results on the task completion and then demonstrate that our proposed framework consistently outperforms SOTA supervised personalization models for the personalization task.

\subsection{Quantitative Analysis}
\label{results-quantitative}
\stitle{Task Completion.}
Despite the fact that the core task in this work is personalization, the personalized models should not compromise the accuracy of task completion for adapting their behaviors for the profiles of the users. 
Keeping it in mind, we report the results for task completion in Tables~\ref{tab:f1} that presents F1 scores for all five tasks for all the competing models.
In terms of task completion, all the models show competitive performance except MemNN-split. 
The main reason for all the models showing great performance for task completion is that the user never drops out of the conversation, even if the system keeps providing the user with unwanted recommendations or never adapts the linguistic style according to the user. 
Since, the system eventually completes the task (i.e., the user is too patient which is not the case in the real-world), the F1 score is high for all the competing models. 
Though, the margin is not big, the best models are supervised-GPT and {\ourmodel} (i.e., this work). For example, on tasks one and three, the proposed {\ourmodel} performs the best, and on the remaining three tasks, supervised-GPT shows the best performance. 

It is critical to emphasize that the proposed {\ourmodel} was trained using only 20 labeled training examples in phase three, whereas the supervised-GPT was trained on the complete training set. 
Moreover, we observe that PToD-0 variant (i.e., that was not trained in phase three) has comparable performance when compared to the SOTA personalization models. Last but not least, the few-shot GPT (that skipped phase two training and used only 20 training examples in phase three) baseline does not show good performance for task five as compared to other models.

\begin{figure}[t!]
  \centering
  \includegraphics[width=0.98\linewidth]{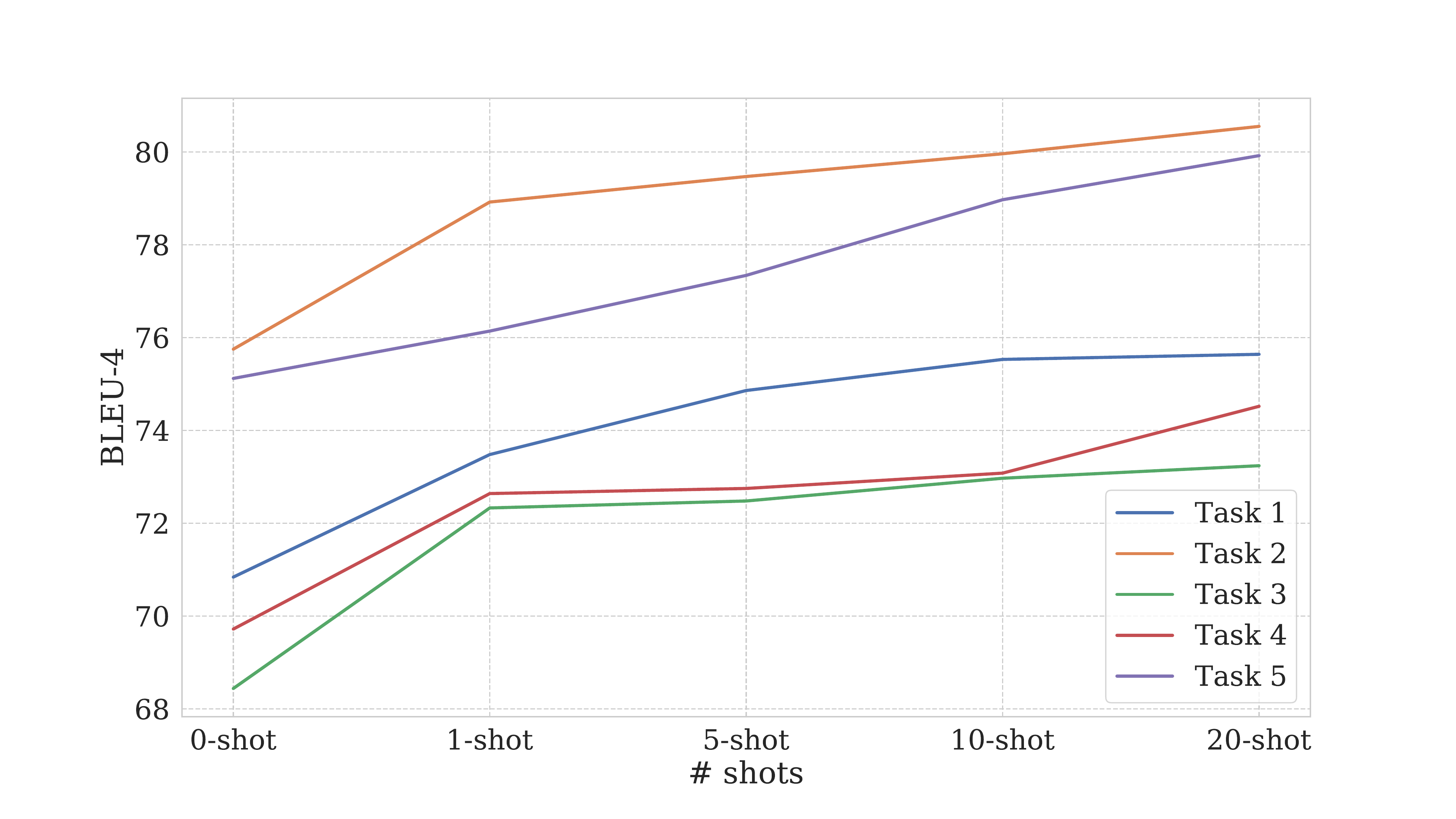}
  \caption{Performance of the {\ourmodel} for different number of shots for all five tasks.
  }
  \label{fig:shots}
  \vspace{-5pt}
\end{figure}

\stitle{Personalization.}
Table~\ref{tab:bleu} presents BLEU-4 and ROUGE-2 scores for all the competing models on all five tasks. For all the tasks, the proposed {\ourmodel} achieves the best performance or insignificant performance difference from supervised-GPT baseline. 
Excluding supervised-GPT model, the proposed {\ourmodel} outperforms all other SOTA response generation methods by at least 19.95\% on BLEU-4 and 9.74\% on ROUGE-2 metrics.
Similarly, the other variant PToD-0 that was not trained on any labeled training examples, still outperforms all the competing models including CoMemNN (which is a response selection model) for BLEU score. 
Since CoMemNN does not generate responses, it has advantage to get better BLEU and ROUGE scores as compared to the response generation approaches. 
Moreover, the few-shot GPT baseline shows the worst performance, since it was trained with only 20 labeled examples in the phase three and phase two (i.e., unsupervised personalization) was skipped. The poor performance of the few-shot GPT baseline highlights the critical role of the phase two.

Figure~\ref{fig:shots} presents the performance of the proposed personalization framework, when provided with different number of training examples in phase three. 
Generally, we notice that as the number of training examples are increased, the performance improves, which highlights the importance of the supervision. However, we noticed that the performance does not get much better beyond 20 examples. That is almost the point, when {\ourmodel} is as good as supervised-GPT model (i.e., trained on full training set).

\begin{figure}[t!]
  \centering
  \includegraphics[width=0.98\linewidth]{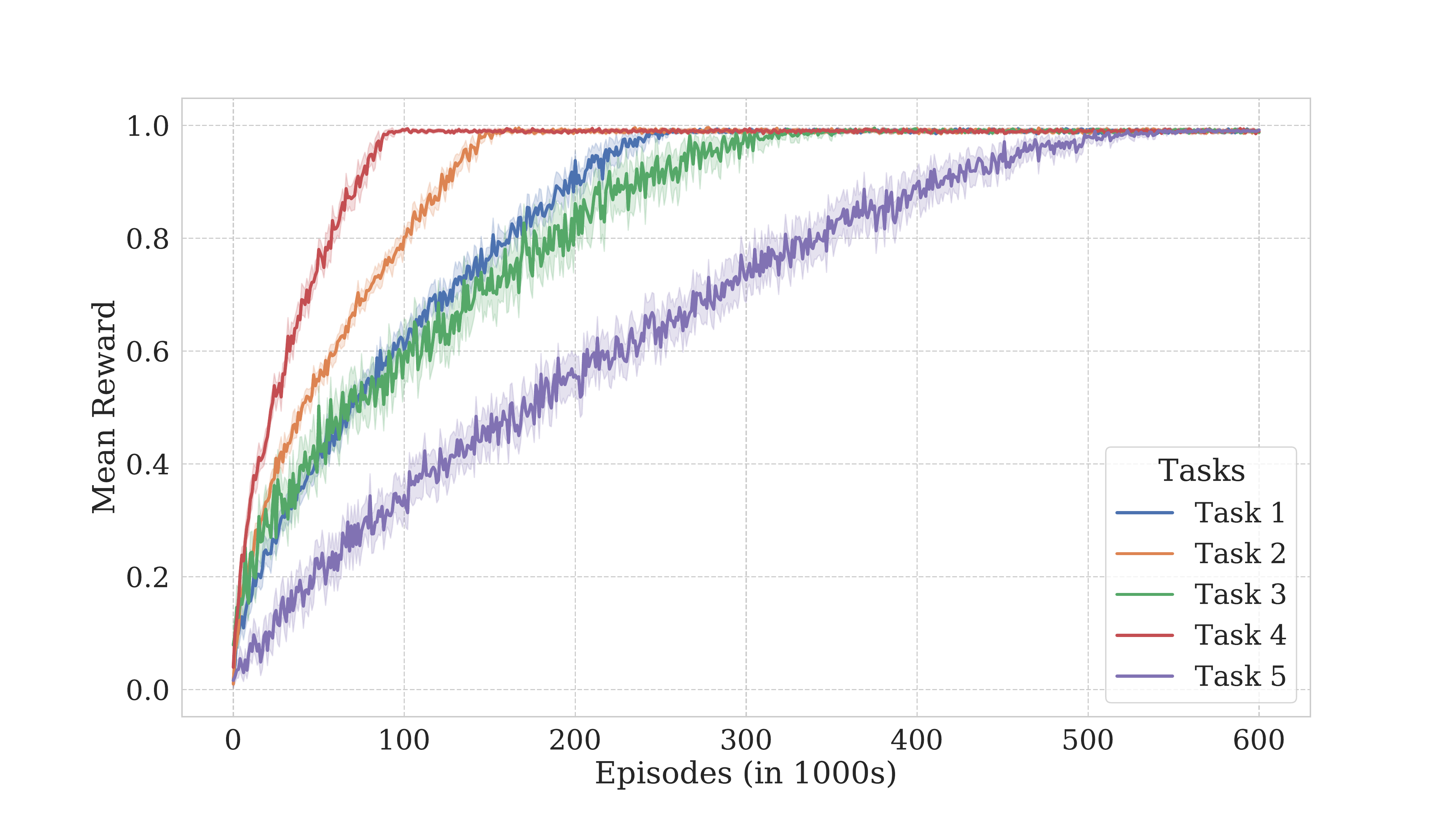}
  \caption{Mean reward across unsupervised personalization phase for all five tasks.
  }
  \label{fig:reward}
  \vspace{-5pt}
\end{figure}

The unsupervised personalization phase is at the core of the proposed framework, we provide more details about it in Figure~\ref{fig:reward}. 
Since all five tasks vary in terms of difficulty, we present the mean reward of the models for each task, as the training progresses in phase two. 
The general trend is that the mean reward starts at 0 (e.g., at episode 0), which is obvious because the responses at the beginning of this phase were not tailored for the given user profile. 
Then, depending on the difficulty of the task, we notice that the respective models start approaching to 1.0 (e.g., after 100,000 episodes). 
We know that the task five (i.e., conduct full personalized dialog) is the most challenging task and the mean reward throughout the training process also signifies that. Similarly, we also notice that the tasks that involve adapting only linguistic styles (e.g., task two), the respective models start to achieve higher mean reward quickly as compared to the tasks that require meaningful recommendations or need to resolve nuances (e.g., task three). 

\begin{table}[t!]
\caption{Average scores of the user study.}
\label{tab:user_study}
\centering
\begin{tabular}{llll}
\toprule
Method         & Fluent & Appropriate & Rank \\
\hline
Reference Response   & 4.92   & 4.87        & 2.41 \\
Supervised GPT & 4.93   & 4.85        & 2.52 \\
PToD-0 (This work)        & 4.91   & 4.86        & 2.62 \\
{\ourmodel}  (This work)          & 4.92   & 4.85        & 2.45 \\
\bottomrule
\end{tabular}
\vspace{-5pt}
\end{table}

\subsection{Qualitative Analysis}
\label{results-qualitative}
In this experiment, we randomly selected 300 responses generated by supervised-GPT (i.e., the best model among the supervised competitors), PToD-0 (i.e., used zero labeled training examples), and {\ourmodel} (i.e., used 20 labeled training examples) along with the reference responses and asked human annotators to rate them (i.e., 1 to 5, 5 being the best) for fluency and appropriateness of the response for the given user profile. 
Moreover, we also asked the annotators to rank the responses for personalization to the given user profile. Each response was rated by three annotators. Table~\ref{tab:user_study} presents average scores for fluency, appropriateness of the response, and average rank among the responses. 
All the models (including reference) achieve high scores on the fluency and appropriateness of the response for the given user profile. 
Moreover, there is not a significant difference among the average scores. Similarly, almost all were ranked similar as reference responses. 
For example, responses generated from every model are ranked at all the places, i.e., $1^{st}$ to $4^{th}$ place. 
In summary, results from human study show that the responses of all the models are as a good as reference responses. 
It is important to remind that the supervised-GPT was trained on the full training set, whereas our proposed PToD-0 and {\ourmodel} were trained using zero and 20 labeled training examples, respectively.

We also observe that the PToD-0 model had slightly lower BLEU and ROUGE scores as compared to {\ourmodel} and supervised-GPT, whereas in the human study it showed equally outstanding performance. Upon further investigation, we noticed that the responses generated by the PToD-0 are identical to that of supervised-GPT and {\ourmodel}. The PToD-0 model did not use the ``words'' (or n-grams) in the reference responses. For example, a perfectly acceptable response generated by PToD-0, ``What should the price be, madam?'' did not get good BLEU or ROUGE scores, because the reference response happened to be, ``Madam! which price range are you looking for?''.  

\section{Related Work}
\label{related}
The two broad categories of dialog systems are open-ended and task-oriented dialog systems. In the following, we summarize the personalization aspect of related work for both categories.

\stitle{Personalized Open-ended Dialogue Systems.}
Among the earlier attempts to personalize open-ended dialog systems, \cite{li2016persona} proposes learning interlocutor persona embeddings and adapting the conversation style accordingly. 
Researchers have since proposed a variety of methods, including persona information fusion~\cite{mazare2018training,zhang2018personalizing}, multi-task learning~\cite{luan2017multi},
transfer learning~\cite{yang2017personalized,zhang2019neural},
meta learning~\cite{madotto2019personalizing},
persona incorporation into the sequence-to-sequence framework~\cite{gulcehre2016pointing,li2016persona},
persona-conditioned RNN-based model~\cite{ficler2017controlling},
persona memory-conditioned variational autoencoders~\cite{song2019exploiting},
response selection using  memory networks~\cite{zhang2018personalizing},
topical information usage~\cite{xu2020neural},
persona pre-training~\cite{herzig2017neural,zheng2020pre}, and
extra training procedures for personalization~\cite{qian2017assigning,herzig2017neural}.
While many of these works have proven useful for assigning personalities or language styles to open-ended dialog systems, they are ineffective for task-oriented dialog systems.
We propose that, rather than assigning personalities to agents (i.e., dialog systems), make them more adaptive to their different kinds of interlocutors in task-oriented dialog settings.

\stitle{Personalized Task-oriented Dialogue
Systems.}
Comparatively to open-domain dialog systems, personalized task-oriented dialog systems are under-explored. In fact, to the best of our knowledge, \myspecial{personalized bAbI dialogue}~\cite{joshi2017personalization} is the only publicly available benchmark for the evaluation of task-oriented dialog systems.
Most of the existing work~\cite{joshi2017personalization,luo2019learning,madotto2018mem2seq,wu2019global,pei2021cooperative} use memory networks by concatenating profile information and dialog memory~\cite{joshi2017personalization}, combining style information~\cite{luo2019learning}, query generation via RNN-based decoder~\cite{madotto2018mem2seq}, local and global encoders~\cite{wu2019global}. 
Similarly, cooperative memory network have been proposed~\cite{pei2021cooperative} to handle the case, where only partial profile information is available. 
All of these works follow supervised learning approaches and require a large amount of labeled training data for each user profile. 
In contrast to previous work, we employ deep reinforcement learning to personalize task-oriented dialog systems in the unsupervised setting without requiring any labeled training data. 
This work leverages pre-trained language models and zero-shot learning for natural language understanding and generation, and adapts its responses to a wide range of user profiles in unsupervised way.
Nonetheless, it is noteworthy to mention that several key ideas leveraged in this work have been used for task-oriented dialog systems such as deep reinforcement learning for dialog policy generation~\cite{le2021predictable,le2021generating} and paraphrasing~\cite{siddique2020unsupervised}, zero-shot learning for
intent detection~\cite{siddique2021generalized} and slot filling~\cite{siddique2021linguistically}, and language models for anaphora resolution~\cite{maqbool2022zero} and response generation~\cite{farooq2020app}. However, none of these works have proposed to personalizing dialog systems in the unsupervised setting.

\section{Conclusion}
\label{conclusion}
We have presented a novel personalization framework for task-oriented dialog systems, {\ourmodel}, that can seamlessly adapt to newly emerging unseen user profiles in the unsupervised fashion.
{\ourmodel} stands out as the first unsupervised framework for personalized task-oriented dialog systems that can effectively adapt its conversation flows and linguistic styles, disambiguate nuances, and make meaningful recommendations according to the profile of the active user.
The key idea behind the proposed framework is using a novel zero-shot generalizable reward function that guides the policy of the personalized model to adapt its responses for the given user without compromising the task completion accuracy.
Our experimental evaluation uses up to 180 diverse user profiles for five tasks including conducting full personalized dialogs. 
Interestingly, our proposed framework outperforms all the existing personalization models using quantitative as well as qualitative analysis. Furthermore, we also trained  a fully supervised-GPT model for comparison and it turned out that {\ourmodel}, trained using only 20 labeled training examples, achieves better or competitive performance.
%\balance

\bibliographystyle{ACM-Reference-Format}
\balance
\bibliography{sample-base}

\end{document}